\documentclass[runningheads]{llncs}
\usepackage{graphicx}
\usepackage{subcaption}
\usepackage{cite}
\usepackage{acro}
\usepackage{float}
\usepackage{xcolor}
\usepackage{amsmath}
\usepackage{multirow}
\usepackage{multicol}
\usepackage{wrapfig}
\usepackage{booktabs}   
\usepackage{subcaption}
\usepackage{wrapfig}
\usepackage{float}
\usepackage[english]{babel}
\usepackage{adjustbox}
\usepackage{amssymb}
\begin{document}
\title{Region-based Contrastive Pretraining for Medical Image Retrieval with Anatomic Query}
\titlerunning{RegionMIR}
% If the paper title is too long for the running head, you can set
% an abbreviated paper title here
%
\author{Ho Hin Lee\inst{1,\thanks{Work conducted during Ho Hin’s internship at Microsoft Health AI.}} \and Alberto Santamaria-Pang\inst{2,\thanks{Corresponding author: \textbf{alberto.santamariapang@microsoft.com}}}\and Jameson Merkow\inst{2}\and Ozan Oktay\inst{3}\and Fernando Pérez-García\inst{3}\and Javier Alvarez-Valle\inst{3}\and Ivan Tarapov\inst{2}}
% \author{Ho Hin Lee\inst{1}\and Yucheng Tang\inst{1}\and Riqiang Gao\inst{1} and Shunxing Bao\inst{1}\and James G. Terry\inst{2}\and J. Jeffrey Carr\inst{2}\and Yuankai Huo\inst{1}\and Bennett A. Landman\inst{1,2}}
%
\authorrunning{Lee et al.}
% First names are abbreviated in the running head.
% If there are more than two authors, 'et al.' is used.

\institute{No Institute Shown}
% \newline
% \email{****@****.****}}
\institute{Vanderbilt University \and Microsoft Health AI \and
Microsoft Health Futures}
% \email{ho.hin.lee@vanderbilt.edu}}
%
\maketitle  % typeset the header of the contribution
\begin{abstract}
We introduce a novel Region-based contrastive pretraining for Medical Image Retrieval (RegionMIR) that demonstrates the feasibility of medical image retrieval with similar anatomical regions. RegionMIR addresses two major challenges for medical image retrieval i) standardization of clinically relevant searching criteria (e.g., anatomical, pathology-based), and ii) localization of anatomical area of interests that are semantically meaningful. Our approach utilizes a Region-Of-Interest (ROI) based image search that works at scale, enabling clinicians to search and retrieve selected ROIs that correspond to the same anatomy and/or similar pathological conditions. Previous approaches match similar images as a whole, without capturing the fine-grained details of specific anatomical regions. In this work, we propose an ROI image retrieval image network that retrieves images with similar anatomy by extracting anatomical features (via bounding boxes) and evaluate similarity between pairwise anatomy-categorized features between the query and the database of images using contrastive learning. ROI queries are encoded using a contrastive-pretrained encoder that was fine-tuned for anatomy classification, which generates an anatomical-specific latent space for region-correlated image retrieval. During retrieval, we compare the anatomically encoded query to find similar features within a feature database generated from training samples, and retrieve images with similar regions from training samples. We evaluate our approach on, both: anatomy classification and image retrieval tasks using the Chest ImaGenome Dataset. Our proposed strategy yields an improvement over state-of-the-art pretraining and co-training strategies, from 92.24 to 94.12 (2.03\%) classification accuracy in anatomies. We qualitatively evaluate the image retrieval performance demonstrating generalizability across multiple anatomies with different morphology.

% \footnote[$\dagger$]{}

\end{abstract}

% \footnote[1]{Work conducted during Ho Hin’s internship at Microsoft Health AI.}
% \footnote[4]{Corresponding author: \textbf{alberto.santamariapang@microsoft.com}}
%
\section{Introduction}
Recent advances in imaging and AI technologies from clinics, hospitals and other medical sites have generated large digital stores of medical data \cite{jimaging7080155}, leading to a number of challenges for robust clinical workflows that provide data security, accessibility, interoperability and retention. However, this trend has also brought with its unprecedented opportunities for exploration of new technical capabilities, such as Content Based Image Retrieval (CBIR) \cite{li2021recent, chen2022deep}. Despite the advances in CIBR for natural images and the increased accessibility of large scale storage for medical images, only limited development for medical image retrieval algorithms has occurred\cite{qayyum2017medical, zhong2021deep, ozturk2021class, choe2022content}. 

A number of challenges obstruct the widespread progress in CBIR for medical imaging including: i) the standardization of a clinical-relevant retrieval criteria (e.g., anatomy, pathology) and ii) the localization of anatomical area of interests that are semantically meaningful. Though recent works have demonstrated feasibility of medical CBIR through regional context with multiple pathologies \cite{zhong2021deep, ozturk2021class, choe2022content,yan2018deeplesion, johnson2019mimic}, these works have been limited to 2D and leverage complete image content. The benefits of robust medical CBIR algorithms to diagnosis and treatment of patients are far reaching. For example, a faster diagnosis and improved treatment plans can be developed through the identification of pathologies by searching for similar characteristics in other patients.  In addition, discovery and recognition of rare diseases through image retrieval are based on specific ROIs within large clinical databases.
\begin{figure}[t!]
    \centering
    \includegraphics[width=\columnwidth]{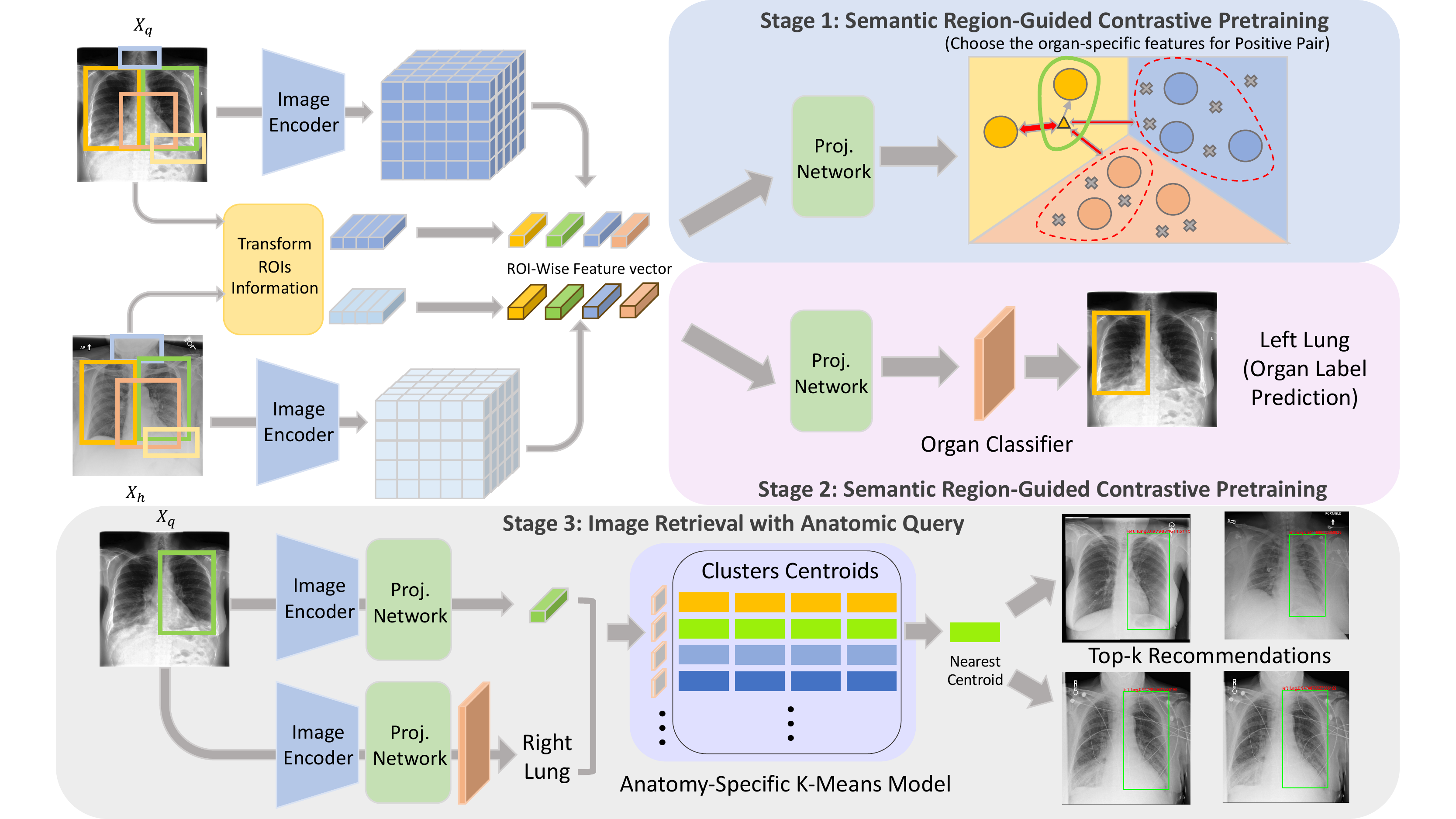}\
    \caption{Overview of RegionMIR. RegionMIR consists of three main steps: 1) regional pooling and constrain the projected vectors into anatomical-specific clusters, 2) fine-tuning both the encoder and the projection network with anatomy classification task to stabilize the anatomical-specific latent space, and 3) training anatomical-specific K-means models to search the most similar centroid for image recommendations.}
    \label{fig:RegionMIR}
\end{figure}
To enhance the efficiency of image retrieval, deep learning frameworks are employed to extract meaningful features and evaluate similarity for retrieval strategy \cite{tang2022deep, hu2022x, choe2022content}. We observed that the image retrieval performance depends on both the quality and the corresponding semantic meanings of the learned representations.
%Contrastive learning provides feasibility to enrich semantic meanings by classifying representations into self-supervised semantic-specific clusters and leads to a significant improvement in downstream tasks \cite{chen2020simple, hin2021semantic, lee2022adaptive, vu2021medaug}.
Contrastive learning offers semantically rich representations by grouping samples into self-supervised semantic-specific clusters, which leads to a significant improvement in downstream tasks \cite{chen2020simple, hin2021semantic, lee2022adaptive, vu2021medaug}. We hypothesize that defining clusters with semantic meanings can enhance the robustness of image retrieval. Currently, most of the contrastive learning strategies involve the use of `whole' images, while limited works are proposed to incorporate both the local and contextual information within region of interests (ROIs). Furthermore, multiple semantic meanings may exist within a subject image or a ROI. It is challenging to define clusters with multiple meanings using current contrastive learning scenarios. As such, we ask: \textit{can we adapt a contrastive learning framework to define regional representations with single semantic meaning and perform robust image retrieval with region queries?}

In this work, we propose a complete hierarchical framework for enhancing medical image retrieval by adapting to anatomy-specific representations in a regional setting. Our framework, named RegionMIR, extends the CBIR task by incorporating regional contrastive learning to generate an anatomy-defined latent space and evaluate the query feature similarity for image retrieval. We employ a methodology built upon BioViL\cite{boecking2022making} and RegionCLIP \cite{zhong2022regionclip}, and further develop a hierarchical search strategy based on the similarity from unsupervised clustering centroids, which differentiates the learned representations into fine-grained conditions (e.g., pathologies) and improves computational efficiency for image retrieval. Our proposed framework is evaluated with one public chest x-ray dataset using 5-fold cross-validation. The experimental results demonstrate a consistent improvement in anatomy classification with a ResNet-50 encoder backbone. Our main contributions are summarized as follows: 1) We propose a hierarchical framework RegionMIR to adapt anatomy-specific representations for image retrieval with regional query; 2) We propose to leverage an unsupervised centroid-based hierarchical search to retrieve top-5 images with reduced time complexity; 3) We demonstrate that RegionMIR learns the region-wise representation with their corresponding semantic meanings, achieving consistent improvements for downstream anatomy classification task and accurate retrieval in different anatomical regions. 

\section{Methods}
%First, we perform ROI-pooling with bounding box query and project the pairwise representations to evaluate the cosine similarity for defining a latent space with anatomical meanings. The contrastive-pretrained network is then fine-tuned with anatomy classification to further differentiate representations into separable clusters. For ROI-based retrieval and support queries in large datasets, we build a hierarchical search strategy based on the similarity from unsupervised clustering centroids. We search the nearest centroid for each query feature and recommend top-5 images with significantly enhanced efficiency (time complexity: $O(n)$ to $O(4 (centroids) * 26 (anatomies))$). 
We introduce the overview of our complete framework RegionMIR in Figure 1. Our primary goal is to learn region-guided representations that are able to differentiate different anatomies within medical images. RegionMIR consists of three hierarchical stages: i) semantic region-guided contrastive pretraining, ii) fine-tuning with anatomy classification, and iii) image retrieval with region query.

\subsection{Semantic Region-Guided Contrastive Pretraining}
Given a set $X=\{(x_1, A_1), (x_2, A_2), \dots, (x_n, A_n)\}$, where $n$ is the total number of images, $A=\{(b_1, y_1), (b_2, y_2),\dots,(b_c, y_c)\}$ is the corresponding bounding box and image-wise labels for all anatomies. The index $c$ represents the total number of semantic regional classes. As a single slice image usually contains rich semantics of multiple anatomies, RegionMIR leverages anatomy-specific bounding boxes to generate large pool of semantic embeddings and learns the regional concepts, regardless of individual full images. First, we randomly sample images from $X$ as query $Q=\{{x_q,A_q}\}_{q=i,…,i+m-1}$ and anchor $H=\{{x_h,A_h}\}_{h=j,…,j+m-1}$, where $i$,$j$ are the randomly sampled index and $m$ is the batch size for training. High-dimensional feature mappings of both query and anchor samples are extracted by the image encoder $E(\cdot)$. Next, we pool features with bounding box label $b$ then linearly project the result into an 1-D embedding space using a multi-layer perceptron (MLP) $P(\cdot)$ as follows:
\begin{equation}
\begin{aligned}
   & \{r_{q,i},\: r_{h,i}\}_{i=1,...,c} = ROIPool(\{E(x_q), b_q\}, \{E(x_h), b_h\}) \\
   & \{z_{q,i},\: z_{h,i}\}_{i=1,...,c} = P(\{r_{q,i},\: r_{h,i}\}_{i=1,...,c}) \\
\end{aligned}
\end{equation}
where $r_{q,i}$ and $r_{h,i}$ are the region-pooled anatomy-specific representations, while $z_{q,i}$ and $z_{h,i}$ are the corresponding projected vectors. Instead of evaluating the image-wise feature similarity between augmented pairs \cite{chen2020simple, khosla2020supervised}, we compute feature similarity region-to-region and define the anatomical-specific features from pairwise samples (query and anchor) as the positive pair. We extend the image-wise contrastive loss to region-wise setting as follows:
\begin{equation}
    \mathcal{L}_{region} = -\frac{1}{c}{\sum_{i=1}^{c}}\sum_{q=1}^{n}\sum_{h=1}^{n} \log \frac{{\exp(\widetilde{z}_{q,i} \cdot \widetilde{z}_{h,i}/\tau)}}{{\sum_{j=0}^{2n-1}}{\sum_{k=1}^{c}} \exp({\widetilde{z}_{q,i} \cdot \widetilde{z}_{j,k}/\tau)}},
    \label{Eq.2} 
\end{equation}
where $z_{j,k}$ are the remaining features within the minibatch. Our proposed regional contrastive loss $\mathcal{L}_{region}$ constrains the same features (possibly with different morphologies) into anatomy-specific embeddings, forcing the model to have a coarse anatomy-to-region correspondence with semantic meanings.

\subsection{Finetuning with Anatomy Classification}
Our ultimate goal is to generate an anatomy-defined latent space and leverages regional features to perform image retrieval of images that contain similar regions. After the encoder network is contrastive-pretrained with a MLP, instead of only fine-tuning the encoder itself due to the image retrieval strategy (discussed in greater details in section 2.3), we fine-tune both the encoder and the MLP followed by a linear layer $L(\cdot)$ for anatomy classification. Finally, we use the multi-class cross-entropy loss to classify each region-pooled feature into specific anatomical embeddings as follows:
\begin{equation}
    \{\tilde{y}_i\}_{i=1,...,c} = L(\{z_{q,i}\}_{i=1,...,c})
\end{equation}
\begin{equation}
    \mathcal{L}_{finetune} = - \frac{1}{|n|}{\sum_{k=0}^{n}} {\sum_{i=0}^{c}} y_{k,i}\log(\tilde{y}_{k,i}),
\end{equation}
by classifying each region-pooled feature with the anatomy label given, the regional representations are further refined and constrained into the anatomical-specific clusters, as shown in Figure \ref{fig:RegionMIR}. With the enriched anatomical meanings in regional representations, such network enhance the feasibility of extending image retrieval with regional similarity and search images with corresponding regional semantics.

\subsection{Image Retrieval with Anatomic Query}
After fine-tuning the learned representations into anatomy-corresponding embeddings, we leverage $P(\cdot)$ project all features from the encoder network into an $L^2$-normalized space and evaluate the feature-wise similarity for image retrieval. Here, we use the cosine similarity to evaluate the representations and retrieve images with the similar region-pooled representations. In our searching scenario, we pre-compute representations from training samples as the retrieval database and use the testing samples as input queries to retrieve images from the database. The most naive approach is to compute similarity between the query representations and all representations in the database for ranking. However, with a time complexity of $O(n)$, this brute force ranking strategy is not computational efficient. Instead, we leverage the anatomy classification predictions and reduce the searching space for each query. Specifically, we train anatomy-specific K-means models to classify the pre-computed representations into different clusters' centroids. We further leverage these anatomy pseudo labels to choose the corresponding K-means model and evaluate the similarity across the centroids. It further enhance the computation efficiency and reduce the time complexity of image retrieval significantly from $O(n)$ to $O(4 (centroids) * 26 (anatomies))$. We finally retrieve the top-$k$ images that are within the highest similarity cluster, where $k$ is the number of searched images.

\section{Implementation Setups}
\textbf{Dataset Details.} In this study, we focus on the Chest ImaGenome dataset \cite{wu2021chest}, which was constructed from the MIMIC-CXR dataset with 2-dimensional chest x-rays \cite{johnson2019mimic}. This dataset consists of two  standards: 1) gold standard and 2) silver standard for dataset curation. Here, we use the gold standard dataset with ground-truth quality label from four independent clinicians to perform training and evaluation. In total, 500 random patients' scans were sampled and annotated with bonding boxes consisting of 26 classes of anatomies (provided in appendix A1) as well as pathology label and clinical attributes.  \\
\textbf{Model Details.} The image encoder is with a ResNet-50 backbone and follows with a projection network, consisting of two consecutive linear layers with ReLU activation in-between. The implementation of training anatomy-specific K-means models is based on scikit-learn and we set the number for unsupervised clusters in each model as 4. Note that the batch size of the anchor input has to be larger than the number of unsupervised clusters, as insufficient samples are limited to demonstrate the distribution of clusters through K-means. \\
\textbf{Training Details.} Both contrastive pre-training and anatomy classification fine-tuning are trained with an Adam optimizer (weight decay: $10^{-4}$; momentum: $0.9$; batch size: 8). For contrastive pre-training, we pre-train the model for 50 epochs with a learning rate of $3 \times 10^{-4}$ and resize all input samples into $512\times512$ without cropping background. For anatomy classification finetuning, we finetune both the encoder and the projection network following with an additional linear layer for an additional 50 epochs a a lower learning rate of $10^{-4}$. During this phase we use the same set of resizing parameters as the pre-training. We perform both training and inference on one Tesla-V100-32GB GPU. Fifty training epochs take about 10 hours with the batch size of 8. \\
\textbf{Performance Comparisons.} We evaluate the quantitative performance in anatomy classification task and the quantitative performance in image retrieval tasks. RegionMIR is compared with different pretraining framework, including BioVIL, and different training scenarios such as co-training downstream task. 
For fair comparison, we report the performance of these methods under 50 epochs finetuning on anatomy classification.

\begin{figure}[t!]
    \centering
    \includegraphics[width=0.95\textwidth]{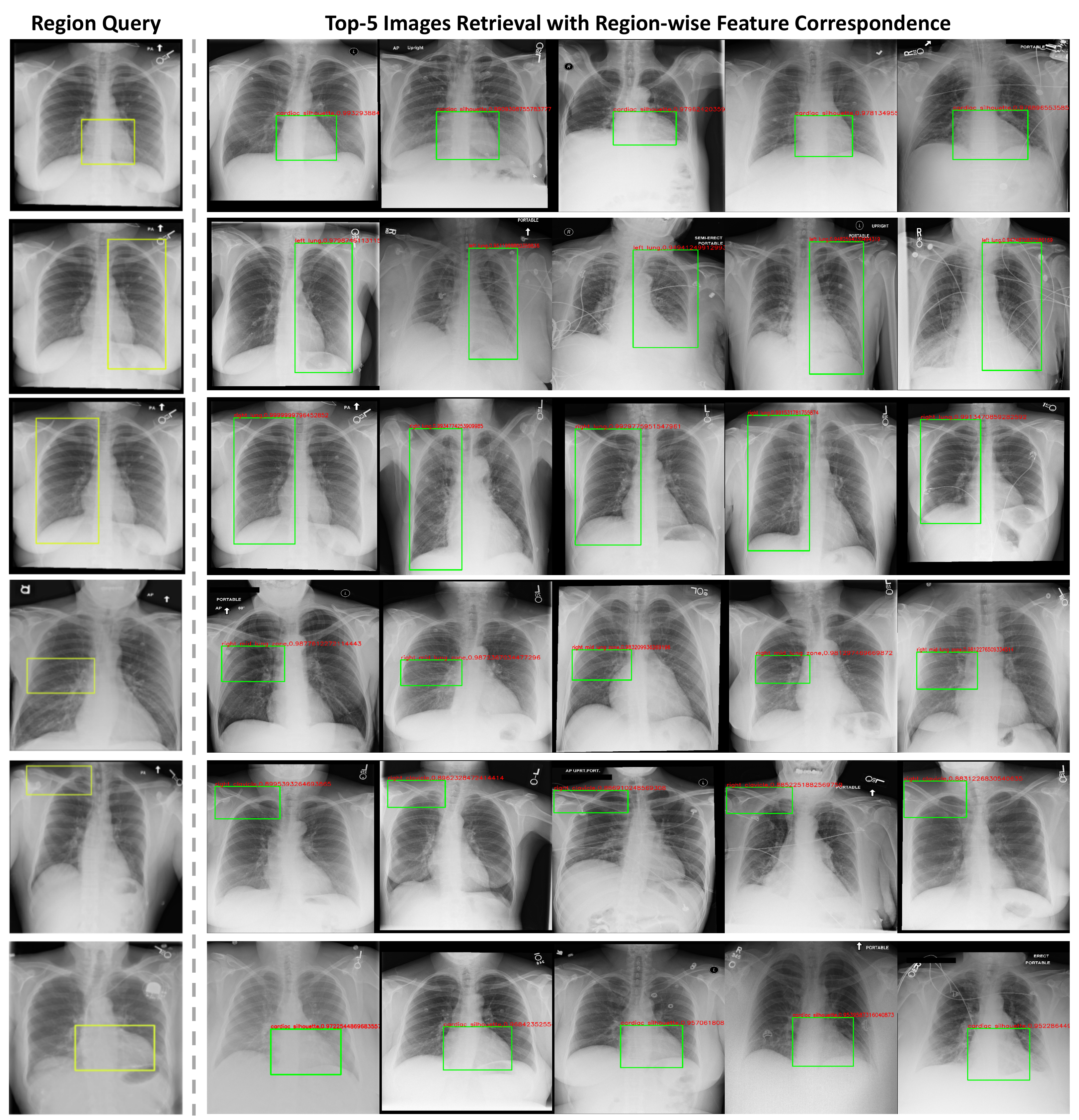}\
    \caption{Top-5 image retrieval examples of different anatomical regions. }
    \label{fig:my_label}
\end{figure}
\section{Results}
\textbf{Anatomy Classification:} From Table 1, we demonstrate the performance quantitatively across different scenarios. By training from scratch, the classification accuracy reaches $92.21$ without contrastive learning strategies. By using transfer learning from BioViL pretrained weights, the classification accuracy significantly improves from $92.21$ to $93.13$. Furthermore, we perform the semantic region-guided contrastive learning in both pretraining and co-training scenarios. Interestingly, the performance further improved from $94.14$ using pretraining scenario, while the performance significantly degraded from $87.94$ using co-training strategies.\\
\textbf{Image Retrieval:} Figure 2, we demonstrate the qualitative representation of image retrieval with multiple anatomic queries. We observe that RegionMIR can successfully retrieve the images with similar anatomical regions. Interestingly, significant morphological differences are demonstrated across the images retrieved. Furthermore, we compute two plots to demonstrates the semantic meanings of the learned representations with our framework (right) and evaluate the similarity between the query representations and the anchor database (left). We found that the learned representations are well separable into clusters with anatomical meanings. For instance, the representations for both left (pink) and right lungs (blue) demonstrate separable clusters with the similarity measures above 0.9, while the clusters with small overlapping regions demonstrate a degradation in similarity measures.
\begin{table}[t!]
    \centering
    \caption{Quantitative Performance on Anatomy Classification}
    \begin{adjustbox}{width=0.85\textwidth}
    \begin{tabular}{*{1}{c}|*{1}{c}|*{1}{c}}
        % \hline \hline
        \toprule
        Pretraining & Finetuning & Classification Accuracy (\%) \\
        \midrule
        $\times$ & $\times$ & $92.24$ \\
        BioVIL & \checkmark & $93.01$ \\
        RegionMIR (Co-train) & $\times$ & $87.94$ \\
        RegionMIR & \checkmark & $94.12$ \\
        \bottomrule
    \end{tabular}
    \end{adjustbox}
    \label{nnUNet_compare}
\end{table}

\begin{figure}[t!]
    \centering
    \includegraphics[width=\textwidth]{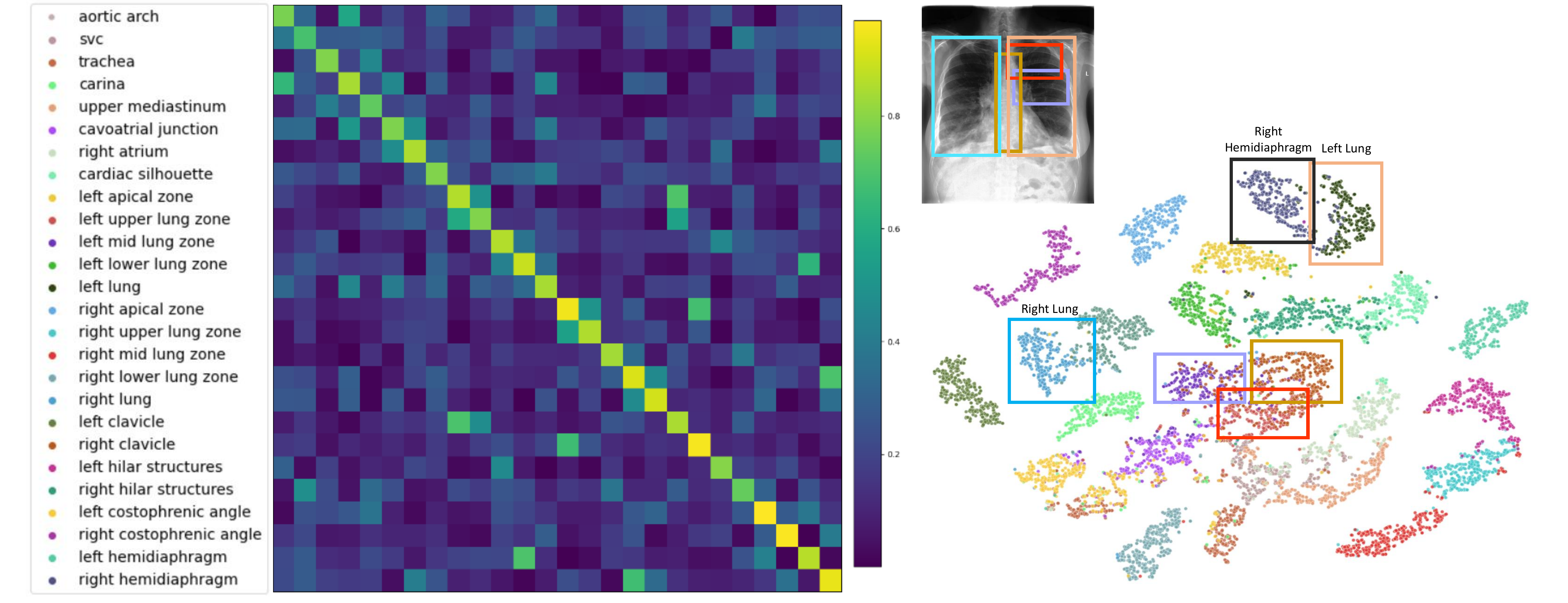}\
    \caption{Left: similarity matrix for anatomical regions of interest, right: TSNE plot with RegionMIR-defined latent space. The order of labels for both the similarity matrix and TSNE plot are aligned. Separable clusters (e.g. left \& right lung) demonstrate high cosine similarity (above 0.9), while clusters with small overlapping region demonstrate a lower cosine similarity value.}
    \label{fig:my_label}
\end{figure}

\section{Discussions}
%In this work, we present a region-based contrastive learning strategy to define regional representations with  anatomical meanings, which extend the feasibility to retrieve image with similar regions. The majority of the current medical image retrieval frameworks demonstrate to extract features conditional to single/multiple labels and perform image retrieval task with whole image setting. However, we observe that multiple conditions (e.g. pathologies, anatomies) may exist in specific ROIs. In our scenario, we extract local representations via ROI pooling and compare the pairwise feature similarity to initially identify semantic-meaningful clusters. 
In figure 1, we have compared the finetuning classification performance by adapting different contrastive pretraining strategies, such as BioVIL. BioVIL adapts multimodal context (e.g. image \& text) for contrastive learning and aligns representations with different semantics, including anatomy and pathologies. We leverage BioVIL as one of our baseline scenario and demonstrate a consistent improvement in classification accuracy (92.24 to 93.01). However, the feature generated from BioVIL is from complete image and the limitation of adapting regional representation still exists in BioVIL. With RegionMIR, the classification performance further improve to 94.12. RegionMIR leverage the bounding box label to specifically pool the local representations and learned the representations with anatomical-meanings alignment. As we define pairwise representations in same anatomy as positive pair, we hypothesize that the latent space is initially defined with anatomy-specific clusters. We further classify representations into clusters with anatomy classification task as model finetuning. Each cluster is well separable with its corresponding anatomical meanings, as shown in Figure 3. For image retrieval scenario, we generate large scale features from training samples as our anchor database, while testing samples are used as the anatomic query. We propose to train K-mean models with anatomy-specific features and search the nearest unsupervised centroid for image recommendations (in Figure 2), which reduce the image retrieval time complexity significantly. The main goal of performing hierarchical search with K-means is to further differentiate the pathology characteristics in the chosen ROI and retrieve images with similar conditions within the ROI. Hierarchical adapting multiple semantic conditions into the contrastive learning framework for region-based image retrieval will be one of our future direction. 

Although RegionMIR have demonstrated the feasibility of adapting regional information for image retrieval, several limitations still exist for the image retrieval task. The first limitation is the dependency on the pseudo prediction of anatomy classification. During the image retrieval, a mismatch K-means model may be chosen due to the inaccurate anatomy prediction, leading to the inaccurate searching criteria and retrieve images with different anatomical focus. Another limitations is the conditional constrains in both contrastive pretraining and the anatomy classification tasks. From Figure 2, we observe that the retrieved image has the corresponding ROI with different morphologies. Such observation may correspond to the definition of positive pair in the contrastive pretraining step. As we randomly sample two images and define their corresponding anatomical feature as the only positive pair, the random sampled image may have significant morphological difference in certain anatomies. Additional constrains for contrastive pretraining will be another potential direction to include multiple constrains for region-based image retrieval.

\section{Conclusion}
We presented RegionMIR, to the best of our knowledge this is the first region-based contrastive learning framework for image retrieval with region query. We extend the image-wise contrastive loss into semantic region-based setting and generate an latent space with anatomy-semantic meanings. Furthermore, we leverage such latent space to retrieve images with similar region query and demonstrate the feasibility of performing region-based image retrieval with significantly reduced time complexity in medical domain. RegionMIR demonstrates a significant improvement on anatomy classification comparing to different pretraining and contrastive learning framework. The latent space of RegionMIR is demonstrated to successfully retrieve images with similar anatomical regions with different morphologies.

\bibliographystyle{splncs04}
\bibliography{MIDL_2023_Microsoft}

\begin{thebibliography}{10}
\providecommand{\url}[1]{\texttt{#1}}
\providecommand{\urlprefix}{URL }
\providecommand{\doi}[1]{https://doi.org/#1}

\bibitem{boecking2022making}
Boecking, B., Usuyama, N., Bannur, S., Castro, D.C., Schwaighofer, A., Hyland,
  S., Wetscherek, M., Naumann, T., Nori, A., Alvarez-Valle, J., et~al.: Making
  the most of text semantics to improve biomedical vision--language processing.
  arXiv preprint arXiv:2204.09817  (2022)

\bibitem{chen2020simple}
Chen, T., Kornblith, S., Norouzi, M., Hinton, G.: A simple framework for
  contrastive learning of visual representations. In: International conference
  on machine learning. pp. 1597--1607. PMLR (2020)

\bibitem{chen2022deep}
Chen, W., Liu, Y., Wang, W., Bakker, E.M., Georgiou, T., Fieguth, P., Liu, L.,
  Lew, M.S.: Deep learning for instance retrieval: A survey. IEEE Transactions
  on Pattern Analysis and Machine Intelligence  (2022)

\bibitem{choe2022content}
Choe, J., Hwang, H.J., Seo, J.B., Lee, S.M., Yun, J., Kim, M.J., Jeong, J.,
  Lee, Y., Jin, K., Park, R., et~al.: Content-based image retrieval by using
  deep learning for interstitial lung disease diagnosis with chest ct.
  Radiology  \textbf{302}(1),  187--197 (2022)

\bibitem{hin2021semantic}
Hin~Lee, H., Tang, Y., Yang, Q., Yu, X., Bao, S., Cai, L.Y., Remedios, L.W.,
  Landman, B.A., Huo, Y.: Semantic-aware contrastive learning for multi-object
  medical image segmentation. arXiv e-prints pp. arXiv--2106 (2021)

\bibitem{hu2022x}
Hu, B., Vasu, B., Hoogs, A.: X-mir: Explainable medical image retrieval. In:
  Proceedings of the IEEE/CVF Winter Conference on Applications of Computer
  Vision. pp. 440--450 (2022)

\bibitem{johnson2019mimic}
Johnson, A.E., Pollard, T.J., Berkowitz, S.J., Greenbaum, N.R., Lungren, M.P.,
  Deng, C.y., Mark, R.G., Horng, S.: Mimic-cxr, a de-identified publicly
  available database of chest radiographs with free-text reports. Scientific
  data  \textbf{6}(1), ~1--8 (2019)

\bibitem{khosla2020supervised}
Khosla, P., Teterwak, P., Wang, C., Sarna, A., Tian, Y., Isola, P., Maschinot,
  A., Liu, C., Krishnan, D.: Supervised contrastive learning. arXiv preprint
  arXiv:2004.11362  (2020)

\bibitem{jimaging7080155}
Kiryati, N., Landau, Y.: Dataset growth in medical image analysis research.
  Journal of Imaging  \textbf{7}(8) (2021). \doi{10.3390/jimaging7080155},
  \url{https://www.mdpi.com/2313-433X/7/8/155}

\bibitem{lee2022adaptive}
Lee, H.H., Tang, Y., Liu, H., Fan, Y., Cai, L.Y., Yang, Q., Yu, X., Bao, S.,
  Huo, Y., Landman, B.A.: Adaptive contrastive learning with dynamic
  correlation for multi-phase organ segmentation. arXiv preprint
  arXiv:2210.08652  (2022)

\bibitem{li2021recent}
Li, X., Yang, J., Ma, J.: Recent developments of content-based image retrieval
  (cbir). Neurocomputing  \textbf{452},  675--689 (2021)

\bibitem{ozturk2021class}
{\"O}zt{\"u}rk, {\c{S}}.: Class-driven content-based medical image retrieval
  using hash codes of deep features. Biomedical Signal Processing and Control
  \textbf{68},  102601 (2021)

\bibitem{qayyum2017medical}
Qayyum, A., Anwar, S.M., Awais, M., Majid, M.: Medical image retrieval using
  deep convolutional neural network. Neurocomputing  \textbf{266},  8--20
  (2017)

\bibitem{tang2022deep}
Tang, Y., Chen, Y., Xiong, S.: Deep semantic ranking hashing based on
  self-attention for medical image retrieval. In: 2022 26th International
  Conference on Pattern Recognition (ICPR). pp. 4960--4966. IEEE (2022)

\bibitem{vu2021medaug}
Vu, Y.N.T., Wang, R., Balachandar, N., Liu, C., Ng, A.Y., Rajpurkar, P.:
  Medaug: Contrastive learning leveraging patient metadata improves
  representations for chest x-ray interpretation. In: Machine Learning for
  Healthcare Conference. pp. 755--769. PMLR (2021)

\bibitem{wu2021chest}
Wu, J.T., Agu, N.N., Lourentzou, I., Sharma, A., Paguio, J.A., Yao, J.S., Dee,
  E.C., Mitchell, W., Kashyap, S., Giovannini, A., et~al.: Chest imagenome
  dataset for clinical reasoning. arXiv preprint arXiv:2108.00316  (2021)

\bibitem{yan2018deeplesion}
Yan, K., Wang, X., Lu, L., Summers, R.M.: Deeplesion: automated mining of
  large-scale lesion annotations and universal lesion detection with deep
  learning. Journal of medical imaging  \textbf{5}(3),  036501 (2018)

\bibitem{zhong2021deep}
Zhong, A., Li, X., Wu, D., Ren, H., Kim, K., Kim, Y., Buch, V., Neumark, N.,
  Bizzo, B., Tak, W.Y., et~al.: Deep metric learning-based image retrieval
  system for chest radiograph and its clinical applications in covid-19.
  Medical Image Analysis  \textbf{70},  101993 (2021)

\bibitem{zhong2022regionclip}
Zhong, Y., Yang, J., Zhang, P., Li, C., Codella, N., Li, L.H., Zhou, L., Dai,
  X., Yuan, L., Li, Y., et~al.: Regionclip: Region-based language-image
  pretraining. In: Proceedings of the IEEE/CVF Conference on Computer Vision
  and Pattern Recognition. pp. 16793--16803 (2022)

\end{thebibliography}

%
% \begin{thebibliography}{8}
% \bibitem{ref_article1}
% Author, F.: Article title. Journal \textbf{2}(5), 99--110 (2016)

% \bibitem{ref_lncs1}
% Author, F., Author, S.: Title of a proceedings paper. In: Editor,
% F., Editor, S. (eds.) CONFERENCE 2016, LNCS, vol. 9999, pp. 1--13.
% Springer, Heidelberg (2016). \doi{10.10007/1234567890}

% \bibitem{ref_book1}
% Author, F., Author, S., Author, T.: Book title. 2nd edn. Publisher,
% Location (1999)

% \bibitem{ref_proc1}
% Author, A.-B.: Contribution title. In: 9th International Proceedings
% on Proceedings, pp. 1--2. Publisher, Location (2010)

% \bibitem{ref_url1}
% LNCS Homepage, \url{http://www.springer.com/lncs}. Last accessed 4
% Oct 2017
% \end{thebibliography}
\end{document}